\documentclass[twoside]{article}
\usepackage[arxiv]{aistats2018}
\newif\ifanonymized
\anonymizedfalse

\usepackage[round]{natbib}
\usepackage[utf8]{inputenc} 
\usepackage[T1]{fontenc}    
\usepackage{lmodern}
\usepackage[hidelinks]{hyperref}       
\usepackage{url}            
\usepackage{booktabs}       
\usepackage{amsfonts}       
\usepackage{nicefrac}       
\usepackage{microtype}      
\usepackage{graphicx}
\usepackage{algorithm}
\usepackage{algpseudocode}
\usepackage{booktabs}
\usepackage{subfigure}
\usepackage{wrapfig}
\usepackage{color}

\usepackage{afterpage}

\usepackage{amsmath}
\usepackage{amssymb}
\usepackage{amsfonts}

\newcommand{\pa}[1]{ \left({#1}\right) }

\def \R {\mathbb{R}}

\def\tbr{{\textbf{r}}}

\def\Te{\textbf{e}}

\def\Tx{\textbf{x}}

\def\Ty{\textbf{y}}

\def\Tz{\textbf{z}}

\def\bsz{\textbf{0}}
\def\bs1{\textbf{1}}
\def\bsa{{\boldsymbol\alpha}}
\def\bse{{\boldsymbol\eta}}

\DeclareMathOperator\mathExp{\mathbb{E}}
\def \E {\mathExp}

\newcommand{\set}[2]{ \left\{ #1 \,\middle|\, #2 \right\} }
\newcommand{\idx}[3]{ \left\{ #1 \right\}_{ #2 }^{ #3 } }

\newcommand{\card}[1]{\left\vert{#1}\right\vert}
\newcommand{\colv}[1]{\begin{pmatrix} #1 \end{pmatrix}}
\newcommand{\mat}[1]{\begin{pmatrix} #1 \end{pmatrix}}

\newcommand{\abs}[1]{\left|#1\right|}

\newcommand{\norm}[1]{\left\| #1 \right\|}

\DeclareMathOperator{\diag}{diag}

\DeclareMathOperator{\Tr}{tr}

\DeclareMathOperator{\cov}{cov}
\DeclareMathOperator{\Unif}{Unif}
\DeclareMathOperator{\blockdiag}{blockdiag}
\DeclareMathOperator{\mvm}{MVM}

\def \mcX {\mathcal{X}}

\def \mcL {\mathcal{L}}

\def\bsa{{\boldsymbol\alpha}}

\def\bse{{\boldsymbol\epsilon}}
\def\bsk{{\boldsymbol\kappa}}

\def\bsth{{\boldsymbol\theta}}

\begin{document}

\ifanonymized

\twocolumn[
\aistatstitle{Large Linear Multi-output Gaussian Process Learning}
\aistatsauthor{ Anonymous Author }
\aistatsaddress{ Anonymous Institution }
]
\else
\twocolumn[
\aistatstitle{Large Linear Multi-output Gaussian Process Learning}
\aistatsauthor{   {Vladimir Feinberg} \And {Li-Fang Cheng} \And  {Kai Li} \And {Barbara E Engelhardt} }
\aistatsaddress{ UC Berkeley \And Princeton University \And Princeton University \And Princeton University }
]
\fi

\begin{abstract}

Gaussian processes (GPs), or distributions over arbitrary functions in a continuous domain, can be generalized to the multi-output case: a linear model of coregionalization (LMC) is one approach \citep{alvarez2012kernels}. LMCs estimate and exploit correlations across the multiple outputs.
While model estimation can be performed efficiently for single-output GPs \citep{msgp}, these assume stationarity, but in the multi-output case the cross-covariance interaction is not stationary.
We propose Large Linear GP (LLGP), which circumvents the need for stationarity by inducing structure in the LMC kernel through a common grid of inputs shared between outputs, enabling optimization of GP hyperparameters for multi-dimensional outputs and low-dimensional inputs.
When applied to synthetic two-dimensional and real time series data, we find our theoretical improvement relative to the current solutions for multi-output GPs is realized with LLGP reducing training time while improving or maintaining predictive mean accuracy. Moreover, by using a direct likelihood approximation rather than a variational one, model confidence estimates are significantly improved.
\end{abstract}

\section{Introduction}\label{introduction}

GPs are a nonlinear regression method that capture function smoothness across inputs through a response covariance function \citep{williams1996gaussian}. GPs extend to multi-output regression, where the objective is to build a probabilistic regression model over vector-valued observations by identifying latent cross-output processes. Multi-output GPs frequently appear in time-series and geostatistical contexts, such as the problem of imputing missing temperature readings for sensors in different locations or missing foreign exchange rates and commodity prices given rates and prices for other goods over time \citep{osborne2008towards, alvarez2010efficient}. Efficient model estimation would enable researchers to quickly explore large spaces of parameterizations to find an appropriate one for their task.

For $n$ input points, exact GP inference requires maintaining an $n^2$  matrix of covariances between response variables at each input and performing $O(n^3)$ inversions with that matrix \citep{williams1996gaussian}. Some single-output GP methods exploit structure in this matrix to reduce runtime \citep{msgp}. In the multi-output setting, the same structure does not exist. Approximations developed for multi-output methods instead reduce the dimensionality of the GP estimation problem from $n$ to $m<n$, but still require $m$ to scale with $n$ to retain accuracy \citep{nguyen2014collaborative}. The polynomial dependence on $m$ is still cubic: the matrix inversion underlying the state-of-the-art multi-output GP estimation method ignores LMC's structure. Namely, the cross-covariance between two outputs is determined by a linear combination of stationary subkernels. On a grid of inputs, each subkernel induces matrix structure, so viewing the LMC kernel as a linear combination of structured matrices we can avoid direct matrix inversion.

Our paper is organized as follows. In Sec.~\ref{sec:background} we give background on single-output and multi-output GPs, as well as some history in exploiting structure for matrix inversions in GPs. Sec.~\ref{sec:related-work} details both related work that was built upon in LLGP and existing methods for multi-output GPs, followed by Sec.~\ref{sec:contributions} describing our contributions. Sec.~\ref{sec:matrix-free} describes our method. Then, in Sec.~\ref{sec:results} we compare the performance of LLGP to existing methods and offer concluding remarks in Sec.~\ref{conclusion}.

\section{Background}
\label{sec:background}

\subsection{Gaussian processes (GPs)}

A GP is a set of random variables (RVs) $\{y_\Tx\}_\Tx$ indexed by $\Tx\in\mcX$, with the property that, for any finite collection $X=\idx{\Tx_i}{i=1}{n}$ of $\mcX$, the RVs are jointly Gaussian with zero mean without loss of generality and a prespecified covariance $K:\mcX^2\rightarrow\R$, $\Ty_X\sim N\pa{\bsz, K_{X, X}}$, where $(\Ty_X)_i=y_{\Tx_i}$ and $(K_{X,X})_{ij}=K(\Tx_i,\Tx_j)$ \citep{williams1996gaussian}. Given observations of $\Ty_X$, inference at a single point $*\in\mcX$ of an $\R$-valued RV $y_*$ is performed by conditioning $y_*|\Ty_X$ \citep{williams1996gaussian}.
Predictive accuracy is sensitive to a particular parameterization of our kernel, and model estimation is performed by maximizing data log likelihood with respect to parameters $\bsth$ of $K$, $\mcL(\bsth)=\log p(\Ty_{X}|X,\bsth)$. Gradient-based optimization methods then require the gradient with respect to every parameter $\theta_j$ of $\bsth$. Fixing $\bsa=K_{X,X}^{-1}\Ty$:

\begin{align}
\partial_{\theta_j}\mcL = \frac{1}{2}\bsa^\top\partial_{\theta_j}K_{X,X}\bsa -\frac{1}{2}\Tr\pa{K_{X,X}^{-1}\partial_{\theta_j}K_{X,X}}.\label{llgradient}
\end{align}

\subsection{Multi-output linear GPs}

We build multi-output GP models as instances of general GPs, where a multi-output model explicitly represents correlations between outputs through a shared input space \citep{alvarez2012kernels}. Here, for $D$ outputs, we write our indexing set as $\mcX'=[D]\times \mcX$, a point from a shared domain coupled with an output index. Then, if we make observations at $X_d\subset\mcX$ for output $d\in[D]$, we can set:
\begin{align*}
X'=\set{(d, \Tx)}{d\in[D],\Tx\in X_d}\subset{\mcX'};\;\; n=\card{X'}.
\end{align*}

An LMC kernel $K$ is of the form:
\begin{align}
K([i,\Tx],[j,\Tz])=\sum_{q=1}^Qb_{ij}^{(q)}k_q(\norm{\Tx-\Tz}),\label{lmcpointwise}
\end{align} 
where $k_q:\R\rightarrow\R$ are stationary kernels on $\mcX$. Typically, the positive semi-definite (PSD) matrices $B_q\in\R^{D\times D}$ formed by $b_{ij}^{(q)}$ are parameterized as $A_qA_q^\top+\diag\bsk_q$, with $A_q\in\R^{D\times R_q},\bsk_q\in\R_+^D$ and $R_q$ a preset rank. Importantly, even though each $k_q$ is stationary, $K$ is only stationary on $\mcX'$ if $B_q$ is Toeplitz. In practice, where we wish to capture covariance across outputs as a $D^2$-dimensional latent process, $B_q$ is not Toeplitz, so $K([i,\Tx],[j,\Tz])\neq K([i+1,\Tx],[j+1,\Tz])$.

The LMC kernel provides a flexible way of specifying multiple additive contributions to the covariance between two inputs for two different outputs. The contribution of the $q$th kernel $k_q$ to the covariance between the $i$th and $j$th outputs is then specified by the multiplicative factor $b_{ij}^{(q)}$. By choosing $B_q$ to have rank $R_q=D$, the corresponding LMC model can have any contribution between two outputs that best fits the data, so long as $B_q$ remains PSD. By reducing the rank $R_q$, the interactions of the outputs have lower-rank latent processes, with rank 0 indicating no interaction (i.e., if $A=0$, then we have an independent GP for each output).

\subsection{Structured covariance matrices}

If we can identify structure in the covariance $K$, then we can develop fast in-memory representations and efficient matrix-vector multiplications (MVMs) for $K$---this has been used in the past to accelerate GP model estimation \citep{gilboa2015scaling, cunningham2008fast}. The Kronecker product $A\otimes B$ of matrices of order $a,b$ is a block matrix of order $ab$ with $ij$th block $A_{ij}B$. We can represent the product by keeping representations of $A$ and $B$ separately, rather than the product. Then, the corresponding MVMs can be computed in time $O(a\mvm(B)+b\mvm(A))$, where $\text{MVM}(\cdot)$ is the runtime of a MVM. For GPs on uniform dimension-$P$ grids, this reduces the runtime of finding $\mcL$ from $O(n^3)$ to $O\pa{Pn^{\nicefrac{1+P}{P}}}$ \citep{gilboa2015scaling}.

Symmetric Toeplitz matrices $T$ are constant along their diagonals and fully determined by their top row $\{T_{1j}\}_{j=1}^n$, yielding an $O(n)$ representation. Such matrices arise naturally when we examine the covariance matrix induced by a stationary kernel $k$ applied to a one-dimensional grid of inputs. Since the difference in adjacent inputs $t_{i+1}-t_{i}$ is the same for all $i$, we have the Toeplitz property that $T_{(i+1)(j+1)}=k(\abs{t_{i+1} -t_{j+1}})=k(\abs{t_i-t_j})=T_{ij}$. Furthermore, we can embed $T$ in the upper-left corner of a circulant matrix $C$ of twice its size, in $O(n\log n)$ time. This approach has been used for fast inference in single-output GP time series with uniformly spaced inputs \citep{cunningham2008fast}. When applied to grids of more than one dimension, the resulting covariance becomes block-Toeplitz with Toeplitz blocks (BTTB) \citep{msgp}. Consider a two-dimensional $n_x\times n_y$ grid with separations $\Delta_x,\Delta_y$. For a fixed pair of $x_1,x_2$, this grid contains a one-dimensional subgrid over varying $y$ values. The pairwise covariances for a stationary kernel in this subgrid also exhibit Toeplitz structure, since we still have $(x_1-x_2,y_i-y_j)=(x_1-x_2,y_i+\Delta_y-y_{j}-\Delta_y)=(x_1-x_2,y_{i+1}-y_{j+1})$. Ordering points in lexicographic order, so that the covariance matrix $K$ has $n_x^2$ order-$n_y$ blocks $K_{x_ix_j}$ with pairwise covariances between varying $y$ values between a pair of fixed $x$ values, the previous sentence implies $\{K_{x_ix_j}\}_{ij}$ are Toeplitz. By similar reasoning, since for any $y_1,y_2$, $(x_i-x_j,y_1-y_2)=(x_{i+1}-x_{j+1},y_1-y_2)$, we have $K_{x_ix_j}=K_{x_{i+1}x_{j+1}}$, thus the block structure of $K$ is itself Toeplitz, and hence $K$ is BTTB (Fig.~\ref{fig:bttb}). For higher dimensions, one can imagine a recursive block-Toeplitz structure. BTTB matrices themselves admit $O(n\log n)$ runtime for MVMs, with $n$ being the total grid size, or $n_xn_y$ in the two dimensional case, and if they are symmetric they can be represented with only their top row as well. The MVM runtime constant scales exponentially in the input dimension, however, so this approach is only applicable to low-input-dimension problems.

\begin{figure}[!ht]
\begin{center}
$$
\begin{array}{c}
K_{i\Delta_x\times Y,j\Delta_x\times Y}=\mat{{\color{red} k\pa{\abs{i-j}\Delta_x} }&{\color{blue}\ddots} \\
{\color{blue}\ddots} &{ \color{red} k\pa{\abs{i-j}\Delta_x}} } \\
K_U=\mat{
 {\color{red} K_{0\times Y} } & {\color{blue} K_{0\times Y,\Delta_x\times Y}}& {\color{green}\ddots }\\
{\color{blue} K_{\Delta_x\times Y, 0\times Y}} & {\color{red} K_{\Delta_x\times Y}} & {\color{blue} K_{\Delta_x\times Y, 2\Delta_x\times Y}} \\
{\color{green} \ddots } & {\color{blue}  K_{2\Delta_x\times Y, \Delta_x\times Y}} & {\color{red} K_{2\Delta_x\times Y}}
}
\end{array}
$$
\end{center}
\caption{An illustration of a stationary kernel $K((a, b), (c, d))=k\pa{\norm{(a-c, b-d)}}$ evaluated on a two-dimensional input grid $U=X\times Y =\{x_i\}\times \{y_j\}$ with separations $\Delta_x,\Delta_y$, resulting in BTTB structure (here, with $\card{X}=3$). Identical matrices are colored the same. We use the shorthand $K_Z=K_{Z,Z}$.}
\label{fig:bttb}
\end{figure}

\section{Related work}
\label{sec:related-work}
\subsection{Approximate inference methods}

Inducing point approaches create a tractable model to approximate the exact GP. For example, the deterministic training conditional (DTC) for a single-output GP fixes inducing points $T\subset\mcX$ and estimates kernel hyperparameters for $\Ty_{X}|\Ty_{T}\sim N(K_{X,T}K_{T,T}^{-1}\Ty_T,\sigma^2)$ \citep{quinonero2005unifying}. This approach is agnostic to kernel stationarity, so one may use inducing points for all outputs $T'\subset \mcX'$, with the model equal to the exact GP model when $T'={X'}$ \citep{alvarez2010efficient}. Computationally, these approaches resemble making rank-$\card{T}$ approximations to the $n\times n$ covariance matrix.

In Collaborative Multi-output Gaussian Processes (COGP), multi-output GP algorithms further share an internal representation of the covariance structure among all outputs at once~\citep{nguyen2014collaborative}. COGP fixes inducing points ${T'}=[D]\times T$ for some $m$-sized $T\subset \mcX$ and puts a GP prior on $\Ty_{{T'}}$ with a restricted LMC kernel that matches the Semiparametric Latent Factor Model (SLFM) \citep{seeger2005semiparametric}. Applying the COGP prior to $\Ty_X$ corresponds to an LMC kernel (Eq.~\ref{lmcpointwise}) where $\bsk_q$ is set to 0 and $A_q=\textbf{a}_q\in\R^{D\times 1}$. Moreover, SLFM and COGP models include an independent GP for each output, represented in LMC as additional kernels $\idx{k_d}{d=1}{D}$, where $A_d=0$ and $\bsk_d=\Te_d\in\R^D$. COGP uses its shared structure to derive efficient expressions for variational inference (VI) for parameter estimation.

\subsection{Structured Kernel Interpolation (SKI)}\label{ski-section}

SKI abandons the inducing-point approach: instead of using an intrinsically sparse model, SKI approximates the original $K_{X,X}$ directly \citep{kiss-gp}. To do this efficiently, SKI relies on the differentiability of $K$. For $\Tx,\Tz$ within a grid $U$, $\card{U}=m$, and $W_{\Tx,U}\in\R^{1\times m}$ as the cubic interpolation weights~\citep{keys1981cubic}, $\abs{K_{\Tx,\Tz}-W_{\Tx,U}K_{U,\Tz}}=O(m^{-3})$. The simultaneous interpolation $W_{X,U}\in\R^{n\times m}$ then yields the SKI approximation: $K_{X,X}\approx W_{X,U}K_{U,U}W_{U,X}^\top$. $W$ has only $4^Pn$ nonzero entries, with $\mcX=\R^P$. Even without relying on structure, SKI reduces the representation of $K_{X,X}$ to an $m$-rank matrix.

Massively Scalable Gaussian Processes (MSGP) exploits structure as well: the kernel $K_{U,U}$ on a grid has Kronecker and Toeplitz matrix structure \citep{msgp}. Drawing on previous work on structured GPs \citep{cunningham2008fast, gilboa2015scaling}, MSGP uses linear conjugate gradient descent as a method for evaluating $K_{X,X}^{-1}\Ty$ efficiently for Eq.~\ref{llgradient}. In addition, an efficient eigendecomposition that carries over to the SKI kernel for the remaining $\log\abs{K_{X,X}}$ term in Eq.~\ref{llgradient} has been noted previously \citep{wilson2014fast}.

Although evaluating $\log\abs{K_{X',X'}}$ is not feasible in the LMC setting because the LMC sum breaks Kronecker and Toeplitz structure, the approach of creating structure with SKI carries over to LLGP.

\section{Contributions of LLGP}\label{sec:contributions}

First, we identify a BTTB structure induced by the LMC kernel evaluated on a grid. Next, we show how an LMC kernel can be decomposed into two block diagonal components, one of which has structure similar to that of SLFM \citep{seeger2005semiparametric}. Both of these structures coordinate for fast matrix-vector multiplication $K\Tz$ with the covariance matrix $K$ for any vector $\Tz$.

With multiple outputs, LMC cross-covariance interactions violate the assumptions of SKI's cubic interpolation, which require full stationarity (invariance to translation in the indexing space $\mcX'$) and differentiability. We show a modification to SKI is compatible with the piecewise-differentiable LMC kernel, which is only invariant to translation along the indexing subspace $\mcX$ (the subkernels $k_q$ are stationary). This partial stationarity induces the previously-identified BTTB structure of the LMC kernel and enables acceleration of GP estimation for non-uniform inputs.

For low-dimensional inputs, the above contributions offer an asymptotic and practical runtime improvement in hyperparameter estimation while also expanding the feasible kernel families to any differentiable LMC kernels, relative to COGP (Tab.~\ref{asymp}) \citep{nguyen2014collaborative}. LLGP is also, to the author's knowledge, first in experimentally validating the viability of SKI for input dimensions higher than one, which addresses a large class of GP applications that stand to benefit from tractable joint-output modeling.

\section{Large Linear GP} \label{sec:matrix-free}

We propose a linear model of coregionalization (LMC) method based on recent structure-based optimizations for GP estimation instead of variational approaches. Critically, the accuracy of the method need not be reduced by keeping the number of interpolation points $m$ low, because its reliance on structure allows better asymptotic performance.
For simplicity, our work focuses on multi-dimensional outputs, one-dimensional inputs, and Gaussian likelihoods.

For a given $\bsth$, we construct an operator $\tilde{K}$ which approximates MVMs with the exact covariance matrix, which for brevity we overload as $K\triangleq K_{X',X'}$, so $K\Tz\approx \tilde{K}\Tz$. Using only the action of MVM with the covariance operator, we derive $\nabla\mcL(\bsth)$. Critically, we cannot access $\mcL$ itself, only $\nabla\mcL$, so we choose AdaDelta as a gradient-only high-level optimization routine for $\mcL$ \citep{zeiler2012adadelta}.

\subsection{Gradient construction}

\citet{gibbs1996cient} describe the algorithm for GP model estimation in terms of only MVMs with the covariance matrix. In particular, we can solve for $\bsa$ satisfying $K\bsa=\Ty$ in Eq.~\ref{llgradient} using linear conjugate gradient descent (LCG). Moreover, Gibbs and MacKay develop a stochastic approximation by introducing RV $\tbr$ with $\cov \tbr=I$:
\begin{align}
  \Tr\pa{K^{-1}\partial_{\theta_j}K} = \E\left[(K^{-1}\tbr)^\top\partial_{\theta_j}K\tbr\right]\label{eq:trace}.
\end{align}
This approximation improves as the size of $K$ increases, so, as in other work \citep{cutajar2016preconditioning}, we let $\tbr\sim\Unif\{\pm 1\}$ and take a fixed number $N$ samples from $\tbr$.

\begin{figure*}[!ht]
\begin{center}
  \subfigure[]{\includegraphics[width=0.4\textwidth]{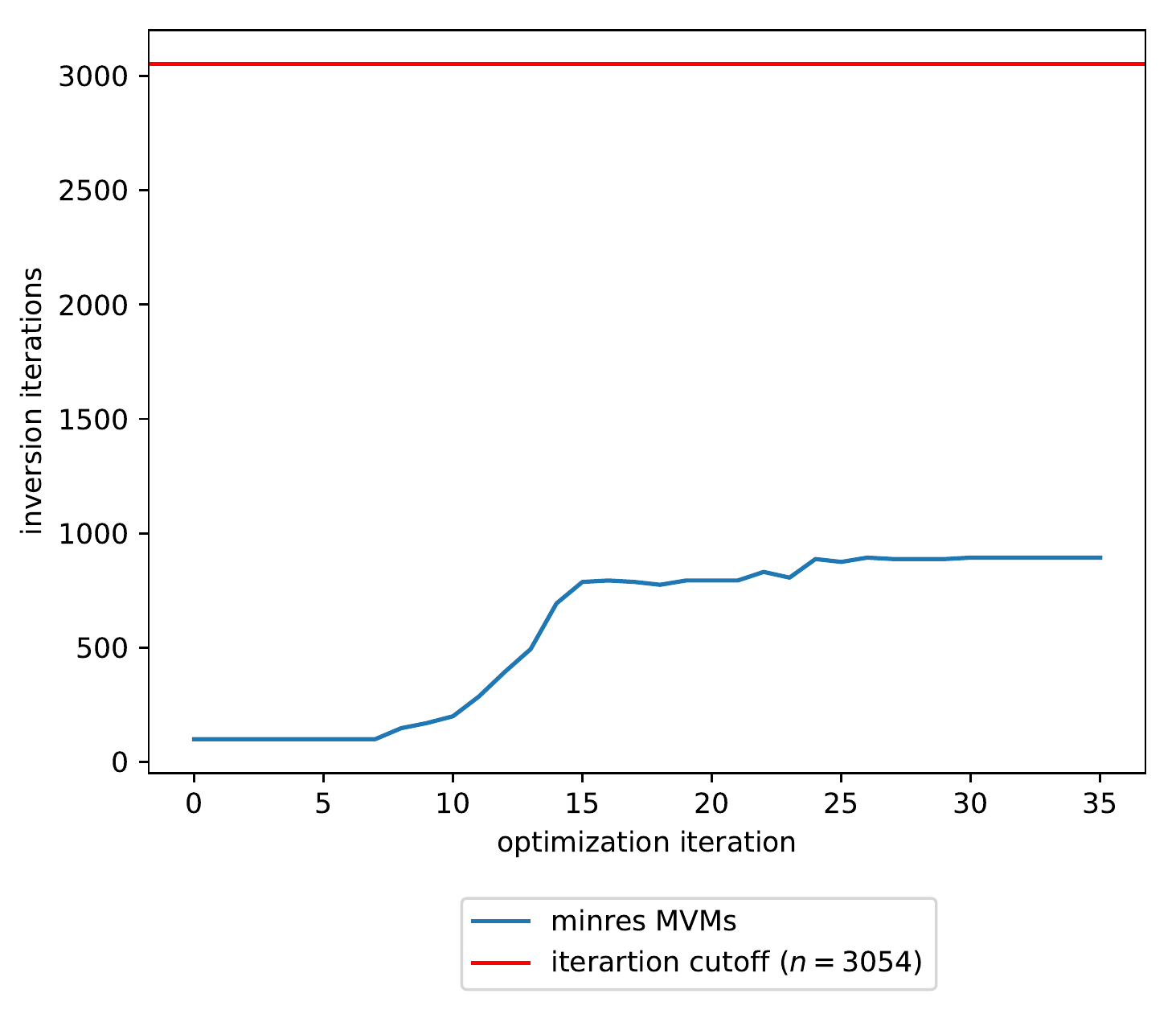}}
  \subfigure[]{\includegraphics[width=0.4\textwidth]{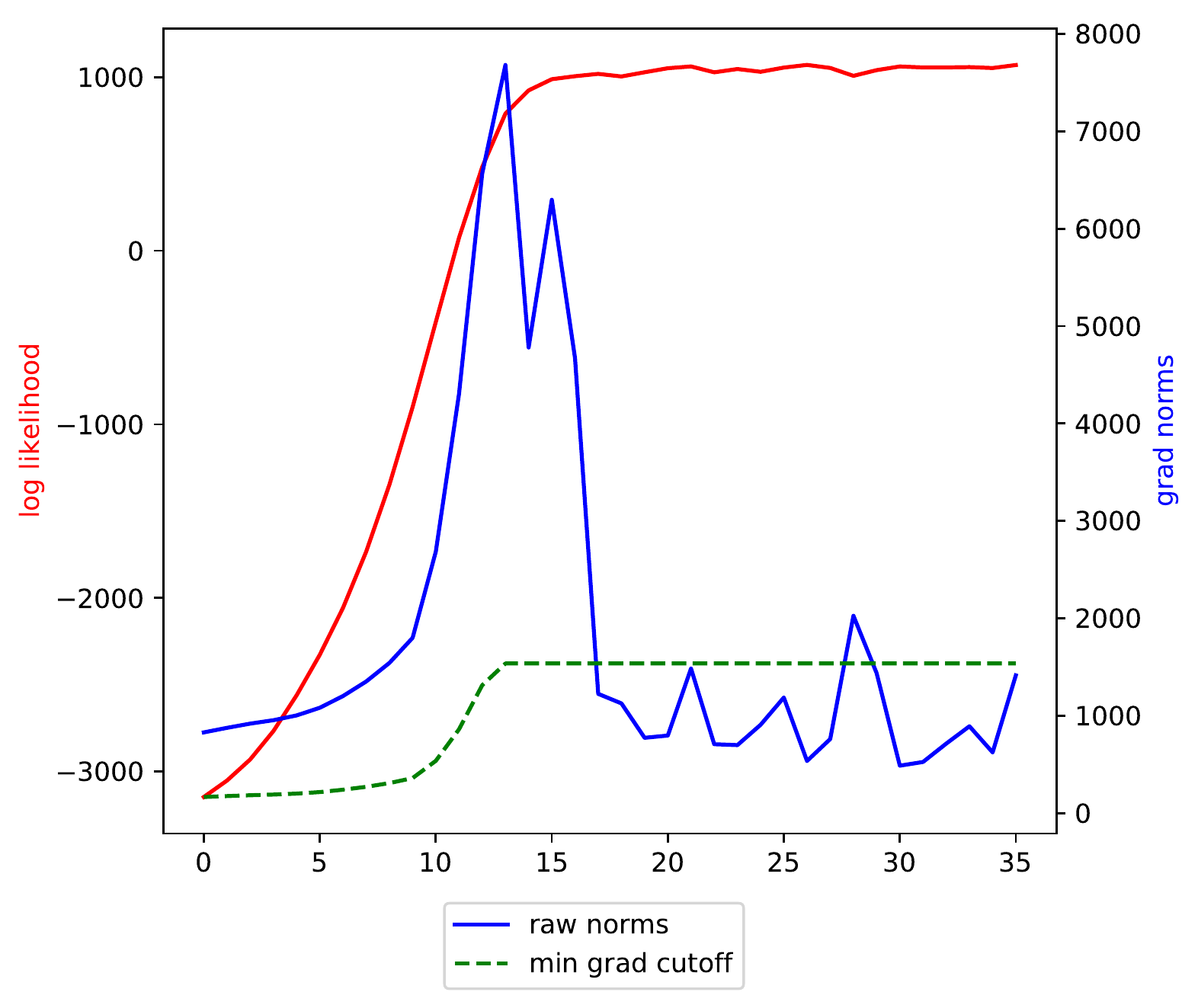}}
\end{center}
\caption{Trace of (a) the number of MVMs that \textsc{minres} must perform to invert $K^{-1}\Ty$ and (b) $\mcL, \norm{\nabla\mcL}$ given $\bsth$ at each optimization iteration for a GP applied to the dataset in Sec.~\ref{fx2007-results}. In (b), in green, we have 20\% of the rolling maximum $\infty$-norm of previous gradients.
}
\label{fx2007-iterations}
\end{figure*}

We depart from Gibbs and MacKay in two important ways (Algorithm~\ref{alg:grad}). First, we do not construct $K$, but instead keep a low-memory representation $\tilde{K}$ (Sec.~\ref{fast-mvm}). Second, we use \textsc{minres} instead of LCG as the Krylov-subspace inversion method used to compute inverses from MVMs. Iterative \textsc{minres} solutions to numerically semidefinite matrices monotonically improve in practice, as opposed to LCG \citep{fong2012cg}. This is essential in GP optimization, where the diagonal noise matrix $\bse$, iid for each output, shrinks throughout learning. Inversion-based methods then require additional iterations because $\kappa_2$, the spectral condition number of $K$, increases as $K$ becomes less diagonally dominant (Fig.~\ref{fx2007-iterations}).

Every AdaDelta iteration (invoking Algorithm~\ref{alg:grad}) then takes total time $\tilde{O}(\mvm(\tilde{K})\sqrt{\kappa_2})$ \citep{raykar2007fast}. This analysis holds as long as the error in the gradients is fixed and we can compute MVMs with the matrix $\partial_{\theta_j}K$ for each $j$ at least as fast as $\mvm({\tilde{K}})$. Indeed, we assume a differentiable kernel and then recall that applying the linear operator $\partial_{\theta_j}$ will maintain the structure of $\tilde{K}$.

For a gradient-only stopping heuristic, we maintain the running maximum gradient $\infty$-norm. If gradient norms drop below a proportion of the running max norm more than a pre-set number of times, we terminate (Fig.~\ref{fx2007-iterations}). This heuristic is avoidable since it is possible to evaluate $\mcL$ with only MVMs \citep{han2015large}, but using the heuristic proved sufficient and results in a simpler gradient-only optimization routine.

\begin{algorithm}[!ht]
  \caption{Compute an approximation of $\nabla \mcL$. Assume \textsc{minres} is the inversion routine. We also assume we have access to linear operators $\partial_{\theta_j}$, representing matrices $\partial_{\theta_j}\tilde{K}$.} \label{alg:grad}
\begin{algorithmic}[1]
  \Procedure{LLGP}{$\tilde{K}$, $\Ty$, $N$, $\{\partial_{\theta_j}\}$}
  \State $R\gets\idx{\tbr_i}{i=1}{N}$, sampling $\tbr\sim\Unif\{\pm 1\}$.
\For{$\Tz$ in $\{\Ty\}\cup R$, in parallel}
\State $K^{-1}\Tz\gets\textsc{minres}(\tilde{K},\Tz)$.
\EndFor
\State $g\gets \bsz$
\For{$\theta_j$ in $\bsth$}\Comment{Compute $\partial_{\theta_j}\mcL$}
\State $t\gets \frac{1}{N}\sum_{i=1}^{N}\pa{K^{-1}\tbr_i}\cdot \partial_{\theta_j}(\tbr_i)$\Comment{Eq.~\ref{eq:trace}}
\State $g_j\gets \frac{1}{2}\pa{K^{-1}\Ty}\cdot \tilde{K}\pa{K^{-1}\Ty}-\frac{1}{2}t$\Comment{Eq.~\ref{llgradient}}
\EndFor
\State \Return $g$ 
\EndProcedure
\end{algorithmic}
\end{algorithm}

\subsection{Fast MVMs and parsimonious kernels}\label{fast-mvm}

The bottleneck of Algorithm~\ref{alg:grad} is the iterative MVM operations in $\textsc{minres}$. Since $K$ only enters computation as an MVM operator, the required memory is dictated by its representation $\tilde{K}$, which need not be dense as long as we can perform MVM with any vector to arbitrary, fixed precision.

When LMC kernels are evaluated on a grid of points for each output, so $X_d=U$, the simultaneous covariance matrix equation without noise over ${U'}$  (Eq.~\ref{lmcgrid}) holds for BTTB matrices $K_q$ formed by the stationary kernels $k_q$ evaluated the shared interpolating points $U$ for all outputs:
\begin{align}
  K_{{U'},{U'}}=\sum_q(A_qA_q^\top+\diag\bsk_q)\otimes K_q\label{lmcgrid}.
\end{align}
To adapt SKI to our context of multi-output GPs, we build a grid ${U'}\subset\mcX'$ out of a common subgrid $U\subset\mcX$ that extends to all outputs with ${U'} = [D]\times U$. Since the LMC kernel evaluated between two sets of outputs $K_{X_i,X_j}$ is differentiable, as long as $U$ spans a range larger than each $\{X_d\}_{d\in[D]}$, the corresponding SKI approximation (Eq.~\ref{lmcski}) holds with the same asymptotic convergence cubic in $1/m$.
\begin{align}
  K\approx \tilde{K}=WK_{{U'},{U'}}W^\top+\bse.\label{lmcski}
\end{align}
We cannot fold the Gaussian noise $\bse$ into the interpolated term $K_{{U'},{U'}}$ since it does not correspond to a differentiable kernel. However, since $\bse$ is diagonal, it is efficient to represent and multiply with. MVM with $\tilde{K}$ requires MVM by the sparse matrices $\bse, W,W^\top$, which all take $O(n)$ space and time.

We consider different representations of $K_{{U'},{U'}}$ (Eq.~\ref{lmcski}) to reduce the memory and runtime requirements for performing the multiplication $K_{{U'},{U'}}\Tz$ in the following sections.

\subsubsection{Sum representation}
 
In \textsc{sum}, we represent $K_{{U'},{U'}}$ with a $Q$-length list. At each index $q$, $B_q$ is a dense matrix of order $D$ and $K_q$ is a BTTB matrix of order $m$, represented using only the top row. In turn, multiplication $K_{{U'},{U'}}\Tz$ is performed by multiplying each matrix in the list with $\Tz$ and summing the results. The Kronecker MVM $(B_q\otimes K_q)\Tz$ may be expressed as $D$ fast BTTB MVMs with $K_q$ and $m$ dense MVMs with $B_q$. In turn, assuming $D\ll m$, the runtime for each of the $Q$ terms is $O(Dm\log m)$.

\subsubsection{Block-Toeplitz representation}

In \textsc{bt}, we note that $K_{{U'},{U'}}$ is a block matrix with blocks $T_{ij}$:
\begin{align*}
\sum_qB_q\otimes K_q =\mat{T_{ij}}_{i,j\in[D]^2},\;\; T_{ij}=\sum_qb_{ij}^{(q)}K_q.
\end{align*}
On a grid $U$, these matrices are BTTB because they are linear combinations of BTTB matrices. \textsc{bt} requires $D^2$ $m$-sized rows to represent each $T_{ij}$. Then, using the usual block matrix multiplication, an MVM $K_{{U'},{U'}}\Tz$ takes $O(D^2m\log m)$ time since each inner block MVM is accelerated due to BTTB structure.

On a grid of inputs with ${X'}={U'}$, the SKI interpolation becomes $W=I$. In this case, using \textsc{bt} alone leads to a faster algorithm---applying the Chan BTTB preconditioner reduces the number of MVMs necessary to find an inverse~\citep{chan1994circulant}.

\subsubsection{SLFM representation}

For the rank-based \textsc{slfm} representation, let $R\triangleq \nicefrac{\sum_qR_q}{Q}$ be the average rank, $R\le D$, and re-write the kernel:
\begin{align*}
  K_{{U'},{U'}}=\sum_q\sum_{r=1}^{R_q}\textbf{a}_q^{(r)}{\textbf{a}_q^{(r)}}^\top\otimes K_q + \sum_q\diag\bsk_q \otimes K_q.
\end{align*}
Note $\textbf{a}_q^{(r)}{\textbf{a}_q^{(r)}}^\top$ is rank $1$. Under some re-indexing $q'\in[RQ]$, which flattens the double summation such that each $q'$ corresponds to a unique $(r, q)$, the term $\sum_q\sum_{r=1}^{R_q}\textbf{a}_q^{(r)}{\textbf{a}_q^{(r)}}^\top\otimes K_q $ may be rewritten as
\begin{align*}
  \sum_{q'}\textbf{a}_{q'}\textbf{a}_{q'}^\top\otimes K_{q'} = \textbf{A}\blockdiag_{q'}\colv{K_{q'}}\textbf{A}^\top,
\end{align*}
where $\textbf{A}=\mat{ \textbf{a}_{q'}}_{q'}\otimes I_m$ with $\mat{ \textbf{a}_{q'}}_{q'}$ a matrix of horizontally stacked column vectors \citep{seeger2005semiparametric}. Next, we rearrange the remaining term $\sum_q\diag\bsk_q \otimes K_q$ as $\blockdiag_d(T_d)$, where $T_d=\sum_q \kappa_{qd}K_q$ is BTTB. Thus, \textsc{slfm} represents $K_{{U'},{U'}}$ as the sum of two block diagonal matrices of block order $QR$ and $D$, where each block is a BTTB order $m$ matrix; thus, MVMs run in $O((QR + D)m\log m)$.

Note that \textsc{bt} and \textsc{slfm} each have a faster run time than the other depending on whether $D^2>QR$. An algorithm that uses this condition to decide between representations can minimize runtime (Tab.~\ref{asymp}). 

\begin{table*}[!ht]
  \caption{
   Asymptotic Runtimes. For both LLGP and COGP, $m$ is a configurable parameter that increases up to $n$ to improve accuracy. $Q,R,D,\kappa_2$ depend on the LMC kernel, which has $O(QRD)$ hyperparameters (Eq.~\ref{lmcpointwise}). The asymptotic performance is given in the table. COGP is only independent of $R$ because it cannot represent models for $R\neq 1$. Computing $\nabla\mcL$ at $\bsth$ requires an up-front cost in addition to the per-hyperparameter cost for each $\theta_j\in\bsth$. Multiplicative log terms in $\kappa_2, m$ are hidden, as are exponential dependencies of the input dimension.
  }
\label{asymp}
\begin{sc}
\begin{center}
\begin{small}
\begin{tabular}{ccc}
  \toprule
  Method & Up-front cost for $\nabla \mcL$ & Additional cost per hyperparameter\\
  \midrule
  Exact & $n^3$ & $n^2 $\\
  COGP & $Qm^3$ & $nm$ \\
  LLGP & $\sqrt{\kappa_2} \pa{n +\min(QR+D, D^2) m}$ &  $n + D m$ \\
  \bottomrule
\end{tabular}
\end{small}
\end{center}
\end{sc}
\end{table*}

\subsection{GP mean and variance prediction}
    
The predictive mean can be computed in $O(1)$ time by $K_{*,X}\bsa\approx W_{*,{U'}}K_{{U'},{U'}}\bsa$ \citep{msgp}.

The full predictive covariance estimate requires finding a new term $K_{*,X}K_{X,X}^{-1}K_{X,*}$. This is done by solving the linear system in a matrix-free manner on-the-fly; in particular, $K_{X,X}^{-1}K_{X,*}$ is computed via \textsc{minres} for every new test point $K_{X,*}$. Over several test points, this process is parallelizable.

\section{Results}
\label{sec:results}
We evaluate the methods on held out data by using standardized mean square error (SMSE) of the test points with the predicted mean, and the negative log predictive density (NLPD) of the Gaussian likelihood of the inferred model. NLPD takes confidence into account, while SMSE only evaluates the mean prediction. In both cases, lower values represent better performance. We evaluated the performance of our representations of the kernel, \textsc{sum}, \textsc{bt}, and \textsc{slfm}, by computing exact gradients using the standard Cholesky algorithm over a variety of different kernels (see the supplement).\footnote{Hyperparameters, data, code, and benchmarking scripts are available in
  \ifanonymized
  \texttt{<anonymous repository>}.
  \else
  \url{https://github.com/vlad17/runlmc}.
  \fi
}

Predictive performance on held-out data with SMSE and NLPD offers an apples-to-apples comparison with COGP, which was itself proposed on the basis of these two metrics. Training data log likelihood would unfairly favor LLGP, which optimizes $\mcL$ directly. Note that predicting a constant value equal to each output's holdout mean results in a baseline SMSE of $1$.

Finally, we attempted to run hyperparameter optimization with both an exact GP and a variational DTC approximation as provided by the GPy package, but both runtime and predictive performance were already an order of magnitude worse than both LLGP and COGP on our smallest dataset from Sec.~\ref{fx2007-results} \citep{gpy2014}.

\subsection{Synthetic dataset}\label{synth-bench}

First, we evaluate raw learning performance on a synthetic dataset. We fix the GP index as $\R^2$ and generate a fixed SLFM RBF kernel with $Q=2$, fixed lengthscales, and covariance hyperparameters $\textbf{a}_1,\textbf{a}_2$. We set the output dimension to be $D=5$, so this synthetic dataset might resemble a geospatial GP model for subterranean mineral density, where the various minerals would be different outputs. Sampling $n\approx 50000$ GP values from the unit square, we hold out approximately $2500$ of them, corresponding to the values for the final output in the upper-right quadrant of the sampling square.

We consider the problem of estimating the pre-fixed GP hyperparameters (starting from randomly-initialized ones) with LLGP and COGP. We evaluate the fit based on imputation performance on the held-out values (Tbl.~\ref{tbl:synth}). For COGP, we use hyperparameter settings applied to the Sarcos dataset from the COGP paper, a dataset of approximately the same size, which has the number of inducing points $m=500$. However, COGP failed to have above-baseline SMSE performance with learned inducing points, even on a range of $m$ up to $5000$ and various learning rates. Using fixed inducing points allowed for moderate improvement in COGP, which we used for comparison. For LLGP, we use a grid of size $m=25\times 25=625$ on the unit square with no learning rate tuning. LLGP was able to estimate more predictive hyperparameter values in about the same amount of time it took COGP to learn significantly worse values in terms of prediction.

\begin{table}[!ht]
\caption{Predictive Performance versus Training Time Tradeoffs on the Synthetic Dataset. We evaluate the learned LLGP model with $m=625$. COGP was evaluated with $m=500$, which were used on a similar-sized dataset from the COGP paper, and increasing $m$ did not improve performance. Since COGP does not provide a terminating condition for its optimization, we also show its performance when permitted to train longer, labelled COGP+. All trials were run 3 times, with parenthesized values representing standard error shown below.}
\label{tbl:synth}
\begin{center}
  \begin{small}
    \begin{sc}
\begin{tabular}{lccc}\toprule
Metric & LLGP & COGP & COGP+\\
\midrule
seconds & $161$ ($3$) & $\textbf{101}$ ($\textbf{0}$) & $1640$ ($0$)\\
SMSE & $\textbf{0.12}$ ($\textbf{0.00}$) & $0.47$ ($0.03$) & $0.15$ ($0.00$)\\
NLPD & $\textbf{0.28}$ ($\textbf{0.00}$) & $21.13$ ($0.82$) & $2.46$ ($0.01$)\\

\bottomrule
\end{tabular}

\end{sc}
\end{small}
\end{center}

\end{table}

\subsection{Foreign exchange rates (FX2007)}\label{fx2007-results}

We replicate the medium-sized dataset from COGP as an application to evaluate LLGP performance. The dataset includes ten foreign currency exchange rates---CAD, EUR, JPY, GBP, CHF, AUD, HKD, NZD, KRW, and MXN---and three precious metals---XAU, XAG, and XPT---implying that $D=13$. In LLGP, we set $Q=1,R=2$, as recommended for LMC models on this dataset~\citep{alvarez2010efficient}. COGP roughly corresponds to the the SLFM model, which has a total of 94 hyperparameters, compared to 53 for LLGP. All kernels are squared exponential. The data used in this example are from 2007, and include $n=3054$ training points and $150$ test points. The test points include 50 contiguous points extracted from each of the CAD, JPY, and AUD exchanges. For this application, LLGP uses $m=n/D=234$ interpolating points. We used the COGP settings from the paper. LLGP outperforms COGP in terms of predictive mean and variance estimation as well as runtime (Tab.~\ref{fx2007-tbl}).

\begin{table}[!ht]
  \caption{Average Predictive Performance and Training Time Over $10$ Runs for LLGP and COGP on the FX2007 Dataset. Parenthesized values are standard error. LLGP was run with LMC set to $Q=1$, $R=2$, and $234$ interpolating points. COGP used a $Q=2$ kernel with $100$ inducing points.}
\label{fx2007-tbl}
\begin{center}
  \begin{small}
    \begin{sc}
      \begin{tabular}{lcc}\toprule
Metric & LLGP & COGP\\
\midrule
seconds & $\textbf{69}$ ($\textbf{8}$) & $96$ ($1$)\\
SMSE & $\textbf{0.21}$ ($\textbf{0.00}$) & $0.26$ ($0.03$)\\
NLPD & $\textbf{-3.62}$ ($\textbf{0.03}$) & $14.52$ ($3.11$)\\

\bottomrule
\end{tabular}

\end{sc}
\end{small}
\end{center}
\end{table}

\begin{figure*}[!ht]
\centering
{\includegraphics[width=\textwidth]{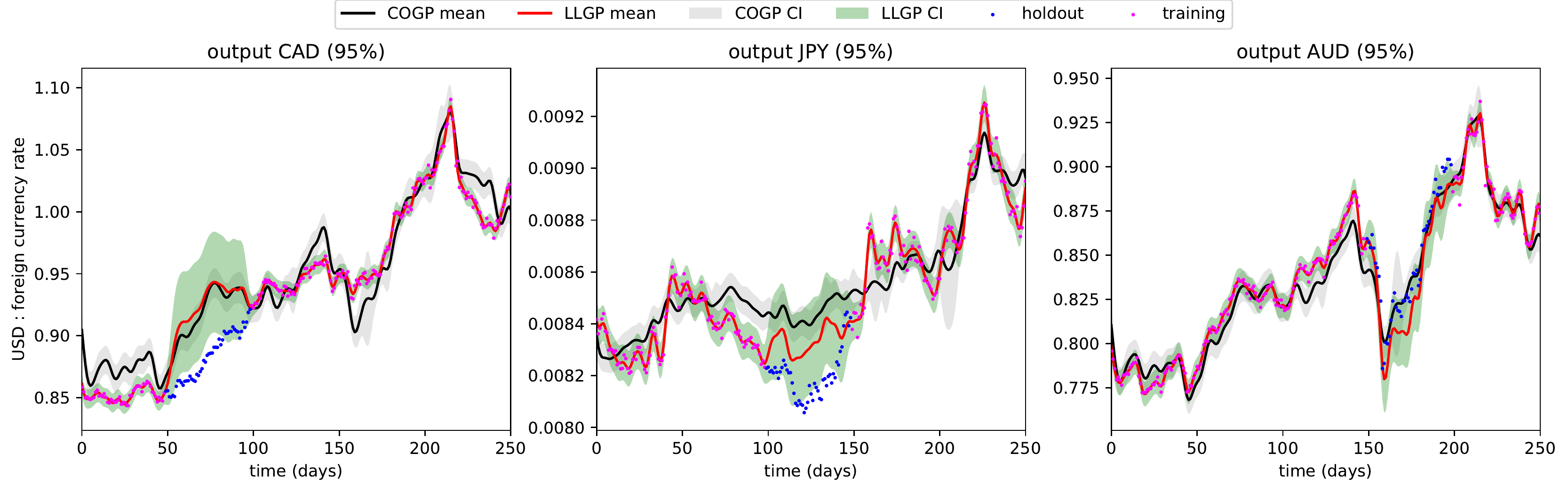}}
\caption{Test outputs for the FX2007 dataset. COGP mean is black, with 95\% confidence intervals shaded in grey. LLGP mean is a solid red curve, with light green 95\% confidence intervals. Magenta points are in the training set, while blue ones are in the test set. Notice LLGP variance corresponds to an appropriate level of uncertainty on the test set and certainty on the training set, as opposed to the uniform variance from COGP.}
\label{fx2007-graph}
\end{figure*}

\subsection{Weather dataset}\label{large-bench}

\begin{figure*}[!ht]
\centering
{\includegraphics[width=\textwidth]{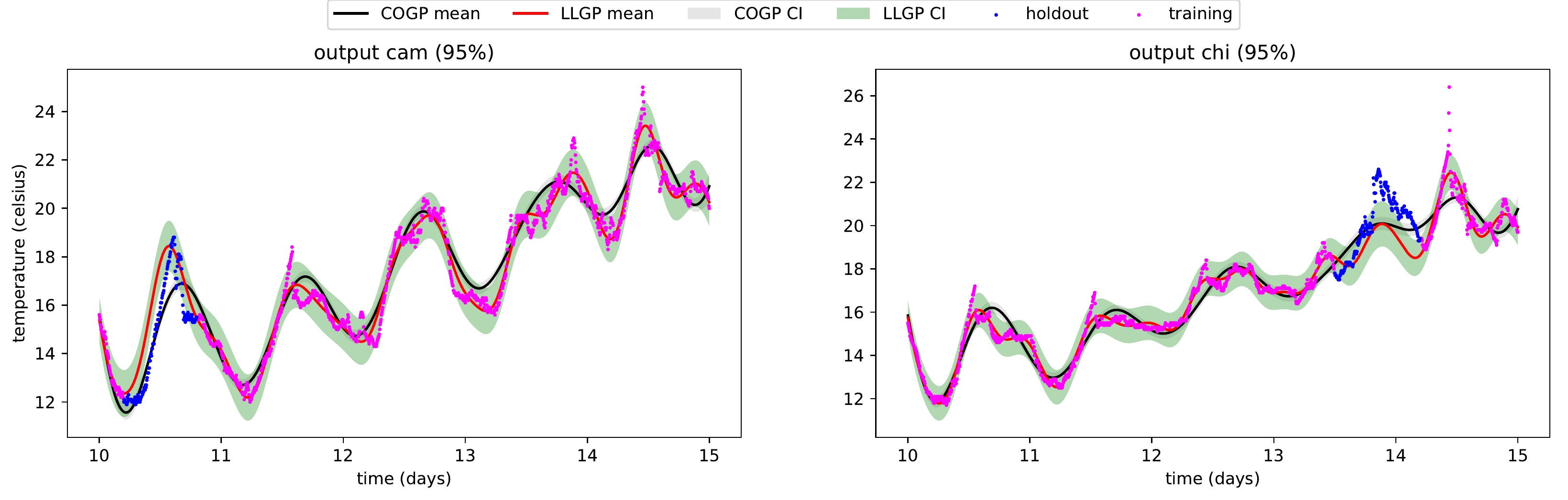}}
\caption{Test outputs for the Weather dataset. COGP mean is black, with 95\% confidence intervals shaded in grey. LLGP mean is a solid red curve, with light green 95\% confidence intervals. Magenta points are in the training set, while blue ones are in the test set. Like Fig.~\ref{fx2007-graph}, the training run was not cherry-picked.}
\label{weather-fig}
\end{figure*}

Next, we replicate results from a weather dataset, a large time series used to validate COGP. Here, $D=4$ weather sensors Bramblemet, Sotonmet, Cambermet, and Chimet record air temperature over five days in five minute intervals, with some dropped records due to equipment failure. Parts of Cambernet and Chimet are dropped for imputation, yielding $n=15789$ training measurements and $374$ test measurements. 

We use the default COGP parameters.\footnote{\url{https://github.com/trungngv/cogp}} We tested LLGP models on $500$ and $1000$ interpolating points.

\begin{table}[!htbp]
  \caption{Average Predictive Performance and Training Time Over $10$ Runs of LLGP and COGP on the Weather Dataset. Parenthesized values are standard error. Both LLGP and COGP trained the SLFM model. We show LLGP with $500$ and $1000$ interpolating points and COGP with $200$ inducing points.}
\label{weather-tbl}
\begin{center}
  \begin{small}
    \begin{sc}
      \begin{tabular}{lccc}\toprule
Metric & \begin{tabular}{c}LLGP\\$m=500$\end{tabular} & \begin{tabular}{c}LLGP\\$m=1000$\end{tabular} & COGP\\
\midrule
seconds & $\textbf{73}$ ($\textbf{12}$) & $90$ ($14$) & $421$ ($4$)\\
SMSE & $0.09$ ($0.01$) & $0.09$ ($0.01$) & $\textbf{0.08}$ ($\textbf{0.00}$)\\
NLPD & $1.72$ ($0.21$) & $\textbf{1.69}$ ($\textbf{0.19}$) & $98.63$ ($1.46$)\\

\bottomrule
\end{tabular}

\end{sc}
\end{small}
\end{center}

\end{table}

LLGP performed slightly worse than COGP in SMSE, but both NLPD and runtime indicate significant improvements (Tab.~\ref{weather-tbl}, Fig.~\ref{weather-fig}). Varying the number of interpolating points $m$ from $500$ to $1000$ demonstrates the runtime versus NLPD tradeoff. While NLPD improvement diminishes as $m$ increases, LLGP still improves upon COGP for a wide range of $m$ by an order of magnitude in runtime and almost two orders of magnitude in NLPD.

\section{Conclusion}\label{conclusion}

In this paper, we present LLGP, which we show adapts and accelerates SKI \citep{kiss-gp} for the problem of multi-output GP regression. LLGP exploits structure unique to LMC kernels,  enabling a parsimonious representation of the covariance matrix, and gradient computations in $\tilde{O}\pa{\sqrt{\kappa_2}(m+n)}$.

LLGP provides an efficient means to approximate the log-likelihood gradients using interpolation. We have shown on several datasets that this can be done in a way that is faster and leads to more accurate results than variational approximations. Because LLGP scales well with increases in $m$, capturing complex interactions in the covariance with an accurate interpolation is cheap, as demonstrated by performance on a variety of datasets (Tab.~\ref{tbl:synth}, Tab.~\ref{fx2007-tbl}, Tab.~\ref{weather-tbl}).

Future work could extend LLGP to accept large input dimensions, though most GP use cases are covered by low-dimensional inputs. Finally, an extension to non-Gaussian noise and use of LLGP as a preconditioner for fine-tuned exact GP models is also feasible in a manner following prior work \citep{cutajar2016preconditioning}.

\pagebreak
\ifanonymized

\else
\subsubsection*{Acknowledgments}
The authors would like to thank Princeton University and University of California, Berkeley, for providing the computational resources necessary to evaluate our method. Further, the authors owe thanks to Professors Joseph Gonzalez, Ion Stoica, Andrew Wilson, and John Cunningham for reviewing drafts of the paper and offering insightful advice about directions to explore.
\fi

\bibliographystyle{unsrtnat}
\bibliography{paper}

\end{document}